# Simplified vision based automatic navigation for wheat harvesting in low income economies

Muhammad Zubair Ahmad, Ayyaz Akhtar, Abdul Qadeer Khan and Amir A. Khan Department of Electrical
Engineering, School of Electrical Engineering and Computer Science
National University of Sciences and Technology, NUST
Islamabad, Pakistan
13mseemahmad@seecs.edu.pk, 13mseeakhtar@seecs.edu.pk, 09beeaqkhan@seecs.edu.pk,
amir.ali@seecs.edu.pk,

**Recent developments in the domain of agricultural robotics have resulted in development of complex and efficient systems. Most of the land owners in the South Asian region are low-income farmers. The agricultural experience for them is still a completely manual process. However, the extreme weather conditions, heat and flooding, often combine to put a lot of stress on these small land owners and the associated labor. In this paper, we propose a prototype for an automated power reaper for the wheat crop. This automated vehicle is navigated using a simple vision based approach employing the low-cost camera and assisted GPS. The mechanical platform is driven by three motors controlled through an interface between the proposed vision algorithm and the electrical drive. The proposed methodology is applied on some real field scenarios to demonstrate the efficiency of the vision based algorithm.**

## I. INTRODUCTION

Agriculture forms backbone of a lot of South Asian economies, a typical case being Pakistan, where agriculture contributes about 21% to the GDP [1]. An important cash crop in the region is wheat. A major handicap in the progress of low-income agricultural economies is the scarcity of labor, caused by rural urban migration. The situation is worsened by severe weather conditions during the harvesting season (mid-April to mid-May) and prolonged exposure to UV radiation have been demonstrated to cause fatigue, cataracts and skin cancer. The monsoon season falling shortly afterwards necessitates exigency of the harvesting process to avoid potential floods. The scale of these natural catastrophes is exacerbated by the lacks of access to proper storage facilities. Considering Pakistan the farmers having landholdings under 5 hectares have an average yield of 2.707 metric tons per hectare [1] which does not translate to a sizable profit due to expensive labor, fuel, fertilizers and pesticides.

   Machine vision, robotics and information technology have seen groundbreaking advances, which has led to development of many cutting edge technologies aiding us in our day to day activities. Agricultural industry can potentially be a huge beneficiary of these developments. More specifically, the focus should be the average farmer, to bring these technological advances to his access at an affordable cost which will go a long way in improving the of global food security situation. The developed solutions must have immediate deploy-ability, with minimal maintenance and possibility of local up-gradation. At the same time, adequate safety measures must be incorporated for a better work environment.

   Agricultural automation was proposed in 1920's [2] but most of the advancements were done after the 1980's. Sensors are critical in this application for the navigation. Different machine vision based guidance algorithms were developed in 1984 [3] and 1985 [4] and implemented on tractors in 1995 [5] and 1997 [6]. Soil-crop segmentation based on Near Infrared (NIR) imaging and Bayes classifier dates back to 1987 [7]. In 1997 [8] a tractor logged 40 hectares based on the cut-uncut crop separation algorithm developed in 1996 [9]. Carrier-phase GPS, real-time Kinematic (RTK) GPS [10] and sensor fusion [11] has also been used for autonomous navigation.

   Tomato picking robot uses edge information and morphological operations to locate the tomato and stereo vision for depth. It has an accuracy of 93.3% in non-overlap condition and 84% otherwise [12]. In radicchio harvesting thresholding in HSL domain and morphological operators are required to mark the center of the plant [13]. In the corn field variable field of view generated by varying pitch and yaw of the camera, has been successfully used to acquire navigation information for the fuzzy controller [14]. Extended kinematic models incorporating slippage [15], cost-effective motion and route planning considering the spatiotemporal constraints of co-operating vehicles [16] has been an important research focus. An example of the later is the relative positioning of combine and transporter so that combine tank does not fill up while the transporter travels a minimum distance. While, the afore mentioned systems are highly effective in their application, they are generally expensive and have been designed for either green crops or for those planted in straight lines (corn, tomato etc.) and thus not well suited to our target crop wheat which is generally planted in a random arrangement. It also has a color similar to soil under bright sunlight. In this paper, we propose an automated vision navigated vehicle to serve as a power reaper for the wheat crop. Our focus will principally be on the overall schematics with special attention to the vision part of the proposed

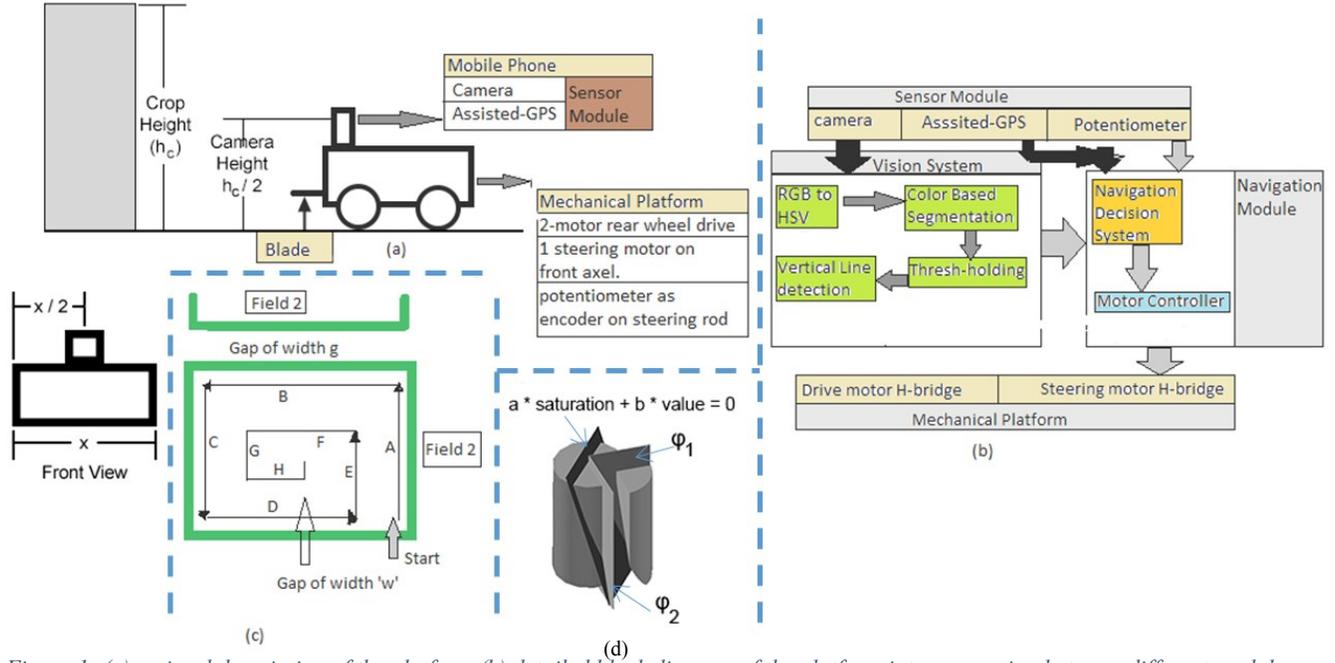

*Figure 1: (a) a visual description of the platform (b) detailed block diagram of the platform interconnection between different modules (c) the model case considered for the control and navigation algorithms (d) the HSV cylinder for subspace decomposition*

scheme. Our proposed method consists of simple sensing mechanism based on mobile phone with built-in camera and assisted-GPS as sensors, while the general purpose processor uses the acquired information to control the actuators (2 rear motors and 1 steering motor) mounted on the vehicle. The control algorithm used is PID with feedback on the degree of turn coming from the potentiometer mounted on the steering rod. The principal focus is on the vision based navigation part, where we demonstrate that a color space transformation from Red Green Blue (RGB) to Hue Saturation Value (HSV) space, segmentation and thresholding in HSV domain and some morphological operations allows independent navigation of a mobile vehicle. The localization within the external boundary is provided by the assisted-GPS.

## II. Proposed Methodology

Fig. 1 illustrates the schematic of the proposed system. The overall system is divided into four modules: sensor module, vision system, navigation system and the mechanical platform. Sensor module consists of the assisted-GPS and the camera, built into the mobile phone. The assisted-GPS has a spatial resolution of 0.5 to 1m, as verified by tracing the boundary of a field in open and then comparing with the physical co-ordinates. The purpose of the assisted-GPS is to acquire the position of the platform with respect to the field to be harvested.

Vision system forms the next module of the integrated system (Fig. 1b). It takes input from the camera in the sensory module and processes it to generate the control signals for navigating and harvesting the field. Based on these signals, the navigation module drives the two rear wheel drive motors and the steering motor to move through the field. The purpose of this vision module is to identify the crops when harvestable (ripe crop still standing), and to control the switching of the blades. The motors and the blades are mounted on the wheeled base, together forming the mechanical part of the vehicle.

Our target crop, wheat, changes color from green to golden-yellow, as it ripens. The surrounding soil has a similar color in bright sunlight due to the luster and the lack of moisture, thus making the contrast between soil and crop indistinguishable. However, if we are able to attenuate the effect of luminosity in the acquired images, this contrast can be enhanced. The acquired images are in RGB color space which is not very efficient for luminance reduction. In this regards, Hue-Saturation-Value (HSV) space is intuitively a better choice as it separates out the effects of luminance and chrominance. Hence, in principle, projection in HSV domain makes it possible to reduce the effects caused by sunlight.

HSV is a cylindrical system ($\rho$, $\theta$, $z$) in which Hue ($\theta$) is a circular scale representing the color variations, Saturation ($\rho$) represents the intensity of the color and Value ($z$) represents the amount of black color mixed into the color. The information of the chrominance, predominantly, lies in the Hue and the Saturation, while the Value contains the information regarding luminance. The first stage of this image processing algorithm is thus transformation from RGB ($Y_{RGB}$) to HSV ($Y_{HSV}$) space. Since, our goal is to enhance the color contrast between the ripe crop and the soil, we create a subspace of this HSV space. This subspace ($SY_{RGB}$) corresponds to the yellow portion of the HSV, as it contains the soil and crop pixels and thus removes the influence of background. This subspace decomposition is expressed in terms of following equations:

$$a * Saturation + b * Value = 0 \qquad (1)$$

$$\Delta\theta = \varphi_2 - \varphi_1, \qquad (2)$$

Where $\varphi_1$ is the first cutting-plane and $\varphi_2$ is the second cutting-plane of hue, and $\Delta\theta$ determines the range of colors which are segmented out. The constants $a$, $b$ vary the saturation and value parameters of the section that is segmented out. Eq. 2 is the section between two half planes and Eq. 1 is a tilted plane which further segments the cylinder. Together these three equations segment out the desired yellow color range which contains the useful information (crop-soil pixels). $\varphi_1, \varphi_2, a$ and $b$ must be tuned manually before the deployment of the platform in the field. These parameters depend on the weather and the on ground conditions so vary from field to field and from time to time. Field tests showed that on the span of a normal day, from 0900 hours to 1600 hours, the tuning must be done about four times. Extreme weather conditions such as overcast conditions and thunderstorms were ruled out in this test, principally because harvesting is conventionally avoided during these times in the south-Asian region.

After the segmentation of this HSV ($Y_{HSV}$) space, the crop-soil contrast is enhanced by mapping the hue values of the extracted yellow range ($SY_{RGB}$) to the whole circular scale. The distinguishability between the crop and soil is now enhanced and they can be separated by thresholding giving us a new subspace ($TSY_{RGB}$). The thresholding parameters must also be tuned before the deployment of the vehicle as they are dependent on the soil, crop, lighting and weather conditions. The process can be automated by maintaining a database of the crop-soil contrast at different time, terrain and lighting conditions. The presence of residual crop (part of harvested stalk still standing in the soil) and the harvested crop (which is dumped into the field to be collected later) calls for a post processing stage to extract the standing crop from the acquired image. In this regard, we exploit the textural properties of the soil and crop. The simplest and the most dominant texture prevalent in the crop is its vertical orientation and in the soil is its horizontal orientation.

Vertical line detection is thus performed on the image after thresholding ($TSY_{RGB}$), with an additional tilt parameter introduced to cater for windy conditions and natural tilt. The value of this tilt is chosen in the range of -5 to +5 degrees. This step allows removal of the already harvested crop (horizontally oriented) but does not remove the residual crop (vertically oriented). This residual crop is removed by introducing a check on length that can be detected in vertical line detection. The residual crop corresponds to the crop that is below the height of cutting blade (Fig. 1a). The check value is thus selected so that the lines below the height of the blade are removed thus removing the regions in the image corresponding to the residue and the environmental noise (dust and straw suspended in air). After applying this process we call our resulting subspace $VTSY_{RGB}$.

Next, consider a typical field model shown in Fig. 1(c). This model is representative of majority of the scenarios faced in harvesting. The model shows two adjacent fields with a separation '$g$ (m)' between them. It shows trajectory (A to H) that would be followed by the vehicle starting from one corner of the field. The vehicle will thus require movement in almost straight direction with 90 degree left turns to choose the other face of the field, traversing it from outside towards the center. Once the outer boundary is cut, the vehicle moves inward by a distance '$w$' which is equal to the width of the blade.

The assisted-GPS marks the co-ordinates of the field before being mounted on the vehicle. The navigation block initially uses the coordinates acquired by the assisted-GPS. For the fields with straight edges, we need to acquire only the corner co-ordinates and then generate the other edges via calculation but for the irregular shaped fields; we need to acquire the coordinates for all the edges. Although, the boundary is clearly marked by the vision system alone but the assisted-GPS is used to make sure that there are no violations of the adjacent field which can lead to legal issues. Another approach is to define this boundary slightly inside our target field to cater for the positional errors of the assisted-GPS. The GPS coordinates are acquired at run time and compared to the previously acquired boundary coordinates. The assisted-GPS can thus aid us in marking the outer boundary along ABCD.

Once, the outer boundary has been harvested, we rely only on the vision based navigation to save computations. In vision based navigation, there are two major tasks: to identify the end of field exactly and to follow the uncut crop. In order to detect the end of field, refer to the Fig.1 (a). As the vehicle reaches the end of path A, there is a gap '$g$' between the fields. The goal is to detect this gap and thus to alert the vehicle about the end of field. Here, we assume that the crop is of same height in both the fields. The laws of optics dictate that if two objects have same physical height, then the pixels of the object that is nearer to the camera appear closer to the top of the image. We draw a line in the image, $l_{height}$, based on a predefined factor e.g., where 80% of the crop pixels lie below it. We fix the $l_{height}$ obtained from first image as the reference line denoted by $l_{reference}$ (Fig. 2 (a)). As long as the vehicle is in Field 1, this line $l_{height}$ will be closer to $l_{reference}$. when all the wheat in the Field 1 has been cut and the wheat of Field 2 appears in the camera, the line $l_{height}$ of the detected crop will be considerably lower than $l_{reference}$, thus forming the cue for the end of field detection. To cater for the height difference in the same field we have a tolerance of 3% of the total pixels around $l_{reference}$. Note that this choice is purely empirical.

To follow the uncut crop we keep the edge of the uncut crop to the right of the vehicle at a certain offset (half of the width of blade =$w/2$) from the center of the vehicle. The right edge of the blade is manually aligned with the outer boundary of the uncut crop at the start of harvesting. From the first acquired image, we mark the position of the outer edge of the crop $E_{outer}$, thus mapping the offset $w/2$ on to the image as $O_{map}$ (Fig. 2 (b)). This ensures that we are navigating the right most uncut crop thus proceeding along the desired path

(ABCD--- and so on) as shown in our model case (Fig.1(c)). If the edge of the uncut crop deviates from this position, the platform turns accordingly. The degree of turn is directly proportional to the deviation. Thus combining the reference line and the offset we have two imaginary lines which help us in the navigation decisions as shown in Fig.2. In case 1: $E_{outer}$ is to the left of $O_{map}$, thus to align them, vehicle moves towards right, in case 2: $E_{outer}$ is to the right of $O_{map}$ so, for alignment the vehicle moves towards left. In case of perfect alignment, the vehicle continues to move straight ahead. The decisions taken by the vision algorithm (i.e. the trajectory and the boundary information) are communicated to the mechanical part of the vehicle via the control interface. The job of this interface will be to control the movement of the vehicle and the switching of the blades.

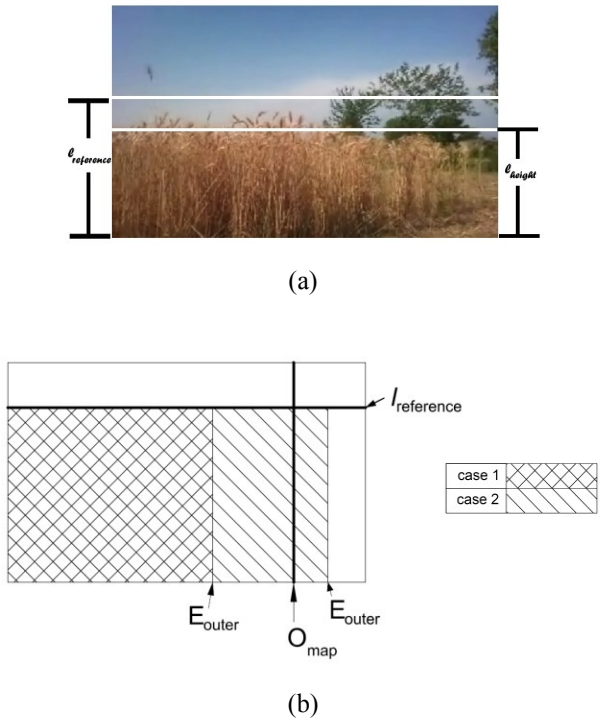

(a)

(b)

Fig 2: Demonstration of vehicle navigation using the imaginary reference line and offset: (a) detection of end of field through the difference between the reference line and the top of the crop in the next field; (b) adjustments in vehicle path depending on the outer crop boundary, case1: $E_{outer}$ is to the left of $O_{map}$, case 2: $E_{outer}$ is to the right of $O_{map}$.

We now focus our attention to the mechanical part of the vehicle. The control algorithm is the direct software interface to the mechanical assembly. There are a total of 3 motors that are controlling the mechanical movement of the vehicle. These motors are controlled via the H-bridges. The two drive motors (rear wheel motors) are controlled by the same H-bridge for coupling reasons while the steering motor by a separate H-bridge. Speed control mechanism is based on Pulse-Width Modulation (PWM). During movement in a straight line, the speed is maintained at a constant value of around 0.5 m/s. The turning command is issued either to follow the uncut crop or at the end of field. For the model case under consideration (Fig 1c), we only need to turn left at every end of field thus simplifying the implementation. The degree of the turn is measured by the potentiometer mounted on the steering rod. This is used as an encoder which implements the steering algorithms in the microcontroller. At the end of field, the 90 degree turn is implemented by maximum turn angle and a PID algorithm. The PID has been tuned in the field tests for each field. The accuracy of the motors can be increased by using hall sensors or optical encoders.

The vision and the navigation system are implemented on the general purpose processor and the control algorithm on the Arduino microcontroller. Both of these communicate via a serial link. The vision algorithms and the navigation algorithms are efficient enough to provide real-time performance.

III. RESULTS AND DISCUSSION

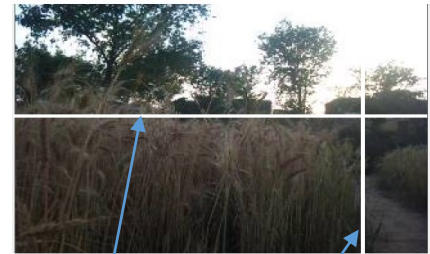

Reference line    $E_{outer}$

(a)

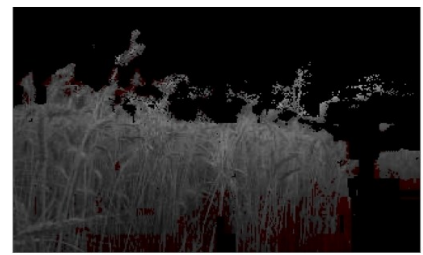

(b)

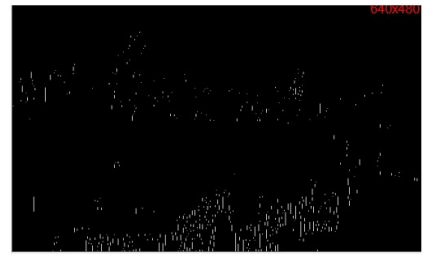

(c)

Fig 3: (a) Original image captured in a field featuring initial placement at the corner of a field. (b) The results after the color-

segmentation and thresholding ($TSY_{HSV}$), the slightly unripe portions are marked red (c) The vertical lines detected in the image

In this section, we demonstrate the results of the proposed vision system and discuss how the navigation is done on the basis of this information. Fig 3(a) displays the initial placement at the corner of the field and we can see homes and trees in the background. Fig 3 (b) shows results after color segmentation and thresholding thus the $TSY_{HSV}$ domain. We can clearly see that the background (trees and the houses) have been removed. This image was taken late in the afternoon so the light was not very bright so the separation between crop and soil was very accurate. The red regions in this image have no navigational significance but mark the slightly unripe portions of the crop. This was done by adjusting the threshold to demonstrate the effectiveness of our method. As this is the first image taken at initial placement the line $l_{height}$ will be set as $l_{reference}$. The placement has been done in such a manner that the outer edge of the crop and the blade are aligned. The outer edge of the crop, $E_{outer}$ as seen in the image (Fig 3a), represents the mapped position of the blade edge, so we will set it as $O_{map}$. This gives us the two references that are used for navigating the vehicle as discussed in the methodology section.

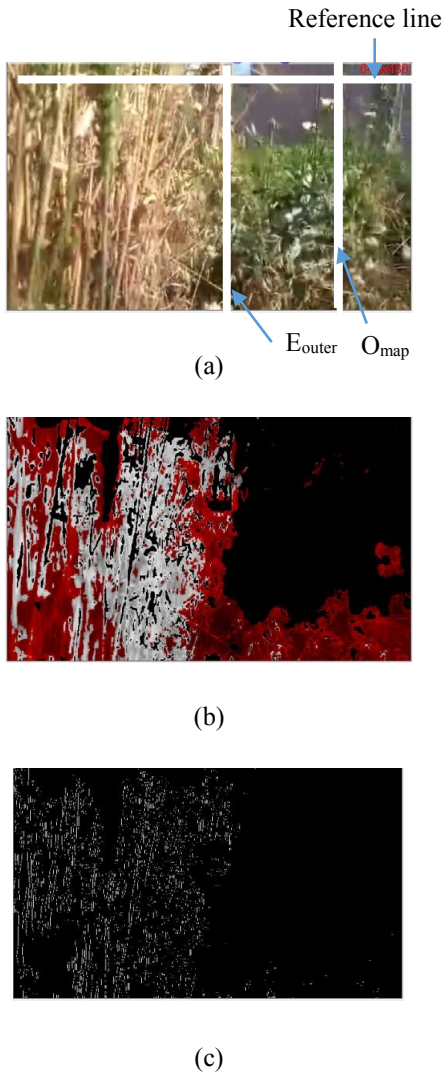

(a)

(b)

(c)

Fig 4: (a) Image showing wheat plants along with weeds (b) The results after the color-segmentation and thresh-holding (c) The vertical lines detected in the image

Fig 4 demonstrates another scenario, in this case the vehicle has drifted somewhat away from the straight line as Fig.4(a) shows the vertical edge has moved to the left of the $O_{map}$, in such a condition the algorithm triggers the steering motor to correct the path by generating appropriate control signal. The current image contains some weeds and some unripe crop plants as well. Fig. 4 (b-c) shows that the algorithm has segregated the ripe crop from the weeds and unripe crop very nicely.

The algorithm was tested under different scenarios, where it offered very satisfactory results. The algorithm requires that some of its parameters are tuned at the start of a harvesting session. These factors can be automatically tuned based on the current time and type of crop, these can be integrated upon future work. Moreover, the purpose of the current work was to develop a proof of concept and considerations for developing an industrial grade product needs further work. In brief, we demonstrated that using a mobile phone camera and assisted-GPS, interfaced to laptop computer, and using a microcontroller to interface the vehicle with the computer, we were able to navigate the vehicle. The main focus of the work was on the development of a simple vision based navigation system. The concept can be extended to industrial scale using embedded systems for computation and industrial scale mechanical design. The control techniques that were not discussed in this paper will have to be adapted for different platforms. We designed and tested the controls for a small autonomous model of a power reaper. We used a model with a steering control and two rear wheel drive motors.

IV. CONCLUSIONS

In this paper, we proposed a simple solution for automated harvesting of wheat crop for low income economies like South East Asia. Wheat, being the popular crop in the region, is usually harvested using manual labor. The climatic conditions being severe in the region require harvesting in a particular time frame. This often leads to labor shortage and necessitates development of low-cost solutions. The paper proposes an algorithm for automated vision based navigation of an agricultural vehicle. We propose a scalable solution as it is independent of the vehicle. The vision algorithm is very simple and computationally efficient for deployment on a real time system. The system is designed to be platform independent and thus can be installed as an add-on component to various existing platforms. The algorithms can be trained for various conditions with a dataset of images under different conditions. The computer used for vision based processing can be replaced by using a dedicated DSP or FPGA, or by utilizing the computational capacities provided by the smart mobile devices. The system can be up-scaled to perform task on a large scale using swarm robotics. Also low cost vehicles can be manufactured to cater for the needs of small scale farmers.

V. REFERENCES

[1] Tech. Report: Economic Survey of Pakistan, 2011

[2] F. Wildrot, "Steering Attachment for Tractors". USA Patent 1506706, 1924.

[3] J.B. Gerrish, T.C Surbrook, "Mobile robots in agriculture," in *First International Conf. on Robotics and Intelligent Machines in Agriculture. ASAE*, St. Joseph, MI, 1984.

[4] J.B.Gerrish, G.C Stockman, "Image processing for path-finding in agricultural field operations," in *ASAE*, St. Joseph, MI, 1985.

[5] B.W. Fehr, J.B. Gerrish, "Vision-guided row crop follower," *Applied Engineering in Agriculture,* vol. 11, no. 4, pp. 613-620, 1995.

[6] J.B. Gerrish, B.W. Fehr, G.R Van Ee, D.P. Welch, "Self-steering tractor guided by computer-vision," *Applied Engineering in Agriculture,* vol. 13, no. 5, pp. 559-563, 1997.

[7] J.F. Ried, S.W. Searcy, "Vision-based guidance of an agricultural tractor," *IEEE Control Systems,* vol. 7, no. 12, pp. 39-43, 1987.

[8] M. Ollis, A. Stentz, "Vision-based perception for an automated harvester," in *IEEE International Conference on Intelligent Robots and Systems*, Piscataway, NJ, 1997.

[9] M. Ollis, A. Stentz, "First results in vision-based crop line tracking," in *IEEE*, Minneapolis, MN, 1996.

[10] M.O'Connor, G. Elkaim, B. Parkinson, "Kinematic GPS for closed-loop control of farm and construction vehicles," in *ION-GPS*, Palm Springs, CA, 1996

[11] N. D Klassen, R.J. Wilson, N.J. Wilson, "Agricultural vehicle guidance sensor," in *ASAE*, St. Joseph, MI, 1993

[12] J. Li, S. Chen, Y. Chen, Y. Chiu, W. Tu, P. Pan, "Study On Machine Vision System For Tomato Picking Robot," in *Proceedings of the 5th International Symposium on Machinery and Mechatronics for Agriculture and Biosystems Engineering (ISMAB*, Fukuoka, Japan, 2010.

[13] M. M. Foglia, G. Reina, "Agricultural Robot for Radicchio Harvesting," Journal of Field Robotics, vol. 23, no. 6, p. 363–377, 2006.

[14] J. Xue, Z. Xhang, T. E. Griffit, "Variable field-of-view machine vision based row guidance of an agricultural robot," *Computer and Electronics in Agriculture,* vol. 84, pp. 85-91, 2012

[15] R. Lenain, B. Thuilot, C. Cariou, P. Martinet, "High accuracy path tracking for vehicles in presence of sliding: Application to farm vehicle automatic guidance for agricultural tasks," *Auton Robot,* vol. 21, pp. 79-97, 2006.

[16] S.Scheuren, S.Stiene, R.Hartanto, J.Hertzberg, M.Reinecke, "Spatio-Temporally Constrained Planning for Cooperative Vehicles in a Harvesting Scenario," *Künstliche Intelligenz,* vol. 27, 2013